\documentclass[runningheads]{llncs}
\usepackage{graphicx}

\usepackage{tikz}
\usepackage{comment}
\usepackage{amsmath,amssymb} 
\usepackage{color}

\usepackage{amsfonts, bm}
\graphicspath{{./figure/}}

\usepackage{dsfont}
\usepackage{gensymb}
\usepackage{multirow}
\usepackage{subcaption}
\usepackage{array}
\newcolumntype{I}{!{\vrule width 1.2pt}}
\newlength\savewidth

\usepackage{booktabs}
\usepackage[
singlelinecheck=false 
]{caption}

\usepackage{setspace}
\begin{document}
\pagestyle{headings}
\mainmatter
\def\ECCVSubNumber{2}  

\title{Investigating Bias and Fairness in Facial Expression Recognition}

\titlerunning{Investigating Bias and Fairness in Facial Expression Recognition}
%
\author{Tian Xu\inst{1} \and
Jennifer White \inst{1} \and
Sinan Kalkan \inst{2} \and
Hatice Gunes \inst{1}}
\authorrunning{T. Xu et al.}
%
\institute{Department of Computer Science and Technology, \\
University of Cambridge, Cambridge, UK \\
\email{\{Tian.Xu,Hatice.Gunes\}@cl.cam.ac.uk, jw2088@cam.ac.uk} \and
Department of Computer Engineering, \\
Middle East Technical University, Ankara, Turkey \\
\email{skalkan@metu.edu.tr}}
\maketitle

\begin{abstract}
Recognition of expressions of emotions and affect from facial images is a well-studied research problem in the fields of affective computing and computer vision with a large number of datasets available containing facial images and corresponding expression labels. However, virtually none of these datasets have been acquired with consideration of fair distribution across the human population. Therefore, in this work, we undertake a systematic investigation of bias and fairness in facial expression recognition
%
by comparing three different approaches, namely a baseline, an attribute-aware and a disentangled approach, on two well-known datasets, RAF-DB and CelebA. Our results indicate that: (i) data augmentation improves the accuracy of the baseline model, but this alone is unable to mitigate the bias effect; (ii) both the attribute-aware and the disentangled approaches equipped with data augmentation perform better than the baseline approach in terms of accuracy and fairness; (iii) the disentangled approach is the best for mitigating demographic bias; and (iv) the bias mitigation strategies are more suitable in the existence of uneven attribute distribution or imbalanced number of subgroup data.

%

\end{abstract}

\section{Introduction}

Automatically recognising expressions and affect from facial images has been widely studied in the literature  \cite{LiDeng-TAC2020,Martinez2019Automatic,Sariyanidi2015Automatic}. 
Thanks to the unprecedented advances in machine learning field, many techniques for tackling this task now use deep learning approaches \cite{LiDeng-TAC2020} which require large datasets of facial images labelled with the expression or affect displayed.

An important limitation of such a data-driven approach to affect recognition is being prone to \emph{bias}es in the datasets against certain demographic groups \cite{bias_face_3,drozdowski2020demographic,bias_face,bias_speech,bias_law,bias_face_2}. The datasets that these algorithms are trained on do not necessarily contain an even distribution of subjects in terms of demographic attributes such as race, gender and age. Moreover, majority of the existing datasets that are made publicly available for research purposes do not contain information regarding these attributes, making it difficult to assess bias, let alone mitigate it. 
Machine learning models, unless explicitly modified, are severely impacted by such biases since they are given more opportunities (more training samples) for optimizing their objectives towards the majority group represented in the dataset.
This leads to lower performances for the minority groups, i.e., subjects represented with less number of samples \cite{bias_face_3,drozdowski2020demographic,bias_face_5,bias_face,bias_speech,bias_law,bias_face_2,bias_face_4}. To address these issues, many solutions have been proposed in the machine learning community over the years, e.g. by addressing the problem at the data level with data generation or sampling approaches \cite{DBLP:conf/aies/AminiSSBR19,Denton2019DetectingBW,iosifidis2018dealing,kamiran2012data,Lu2018GenderBI,NgxandeEtAl2020,DBLP:conf/nips/CalmonWVRV17,wang2020towards}, at the feature level using adversarial learning  \cite{alvi2018turning,morales2019sensitivenets,wang2020towards,zhang2018mitigating} or at the task level using multi-domain/task learning \cite{dwork2012fairness,wang2020towards}.

Bias and mitigation strategies in facial analysis have attracted increasing attention both from the general public and the research communities. For example, many studies have investigated bias and mitigation strategies for face recognition \cite{bias_face_3,bias_face_5,bias_face_11,bias_face,morales2019sensitivenets,bias_face_2,bias_face_4}, gender recognition \cite{bias_face_10,bias_face_5,bias_face_7,bias_face_6}, age estimation \cite{bias_face_8,bias_face_10,bias_face_5,guo2010human,bias_face_7}, kinship verification \cite{bias_face_5} and face image quality estimation \cite{terhorst2020face}. However, bias in facial expression recognition has not been investigated, except for \cite{Denton2019DetectingBW,wang2020towards}, that only focussed on the task of smiling/non-smiling using the CelebA dataset.

In this paper, we undertake a systematic investigation of bias and fairness in facial expression recognition. To this end, we consider three different approaches, namely a baseline deep network, an attribute-aware network  and a representation-disentangling network (following \cite{alvi2018turning,morales2019sensitivenets,wang2020towards,zhang2018mitigating}) under the two conditions of with and without data augmentation. As a proof of concept, we conduct our experiments on RAF-DB and CelebA datasets that contain labels in terms of gender, age and/or race.
%
%
To the best of our knowledge, this is the first work (i) to perform an extensive analysis of bias and fairness for facial expression recognition, beyond the binary classes of smiling / non-smiling \cite{Denton2019DetectingBW,wang2020towards}, (ii) to use the sensitive attribute labels as input to the learning model to address bias, and (iii) to extend the work of \cite{NeurIPS_disent} to the area of facial expression recognition in order to learn fairer representations as a bias mitigation strategy.

\section{Related Work}
Bias in the field of Human-Computer Interaction (HCI), and in particular issues arising from the intersection of gender and HCI have been discussed at length in \cite{BreslinEtAl-2018}. However, studies specifically analysing, evaluating and mitigating race, gender and age biases in affect recognition have been scarce. We therefore provide a summary of related works in other forms of facial analysis including face and gender recognition, and age estimation.

\subsection{Bias and Mitigation in Machine Learning}

Attention to bias and fairness in machine learning (ML) has been rapidly increasing  with the employment of ML applications in everyday life. It is now well accepted that ML models are extremely prone to biases in data \cite{bias_face_3,article1111}, which has raised substantial concern in public such that regulatory actions are being  as preventive measure; e.g. European Commission \cite{european2020white} requires training data for such applications to be "sufficiently broad," and to reflect "all relevant dimensions of gender, ethnicity and other possible grounds of prohibited discrimination". 

Bias mitigation strategies in ML generally take inspiration from data or class balancing approaches in ML, a very related problem which directly pertains to imbalance in the task labels. Bias can be addressed in an ML model in different ways \cite{bellamy2018ai,drozdowski2020demographic}: For example, we can balance the dataset in terms of the demographic groups, using under-sampling or over-sampling \cite{iosifidis2018dealing,wang2020towards}, sample weighting \cite{DBLP:conf/aies/AminiSSBR19,kamiran2012data,DBLP:conf/nips/CalmonWVRV17}, data generation \cite{Denton2019DetectingBW,NgxandeEtAl2020}, data augmentation \cite{Lu2018GenderBI} or directly using a balanced dataset \cite{DBLP:journals/corr/abs-1911-10692,DBLP:conf/iccv/WangDHTH19}. However, it has been shown that balancing samples does not guarantee fairness among demographic groups \cite{WangEtAl-ICCV19}.

Another strategy to mitigate bias is to remove the sensitive information (i.e. gender, ethnicity, age) from the input at the data level (a.k.a. ``fairness through unawareness'') \cite{alvi2018turning,wang2020towards,zhang2018mitigating}. However, it has been shown that the remaining information might be implicitly correlated with the removed sensitive attributes and therefore, the residuals of the sensitive attributes may still hinder fairness in the predictions \cite{dwork2012fairness,hardt2016equality}. Alternatively, we can make the ML model more aware of the sensitive attributes by making predictions independently for each sensitive group (a.k.a. ``fairness through awareness'') \cite{dwork2012fairness}. Formulated as a multi-task learning problem, such approaches allow an ML model to separate decision functions for different sensitive groups and therefore prohibit the learning of a dominant demographic group to negatively impact the learning of another one.Needless to say this comes at a cost - it  dramatically increases the number of parameters to be learned, as a separate network or a branch needs to be learned for each sensitive group.

\subsection{Bias in Facial Affect Recognition}

It has been long known that humans' judgements of facial expressions of emotion are impeded by the ethnicity of the faces judged \cite{kilbride1983ethnic}. In the field of automatic affect recognition, systematic analysis of bias and the investigation of mitigation strategies are still in their infancy. A pioneering study by Howard et al. \cite{HowardEtAl-2017} investigated how using a cloud-based emotion recognition algorithm applied to images associated with a minority class (children'€™s facial expressions) can be skewed when performing facial expression recognition on the data of that minority class. To remedy this, they proposed a hierarchical approach combining outputs from the cloud-based emotion recognition algorithm with a specialized learner. They reported that this methodology can increase the overall recognition results by 17.3\%. %
Rhue \cite{Rhue2018RacialIO}, using a dataset of $400$ NBA player photos, found systematic racial biases in Face++ and Microsoft's Face API. Both systems assigned to African American players more negative emotional scores on average, regardless of how much they smiled.

When creating a new dataset is not straightforward, and/or augmentation is insufficient to balance an existing dataset, Generative Adversarial networks (GAN) have been employed for targeted data augmentation. 
Denton et al. in \cite{Denton2019DetectingBW} present a simple framework for identifying biases in a smiling attribute classifier. They utilise GANs for a controlled manipulation of specific facial characteristics and investigate the effect of this manipulation on the output of a trained  classifier. As a result, they identify which dimensions of variation affect the predictions of a smiling classifier trained on the CelebA dataset. For instance, the smiling classifier is shown to be sensitive to the Young dimensions, and the classification of 7\% of the images change from a smiling to not smiling classification as a result of manipulating the images in this direction. 
Ngxande et al. \cite{NgxandeEtAl2020} introduce an approach to improve driver drowsiness detection for under-represented ethnicity groups by using GAN for targeted data augmentation based on a population bias visualisation strategy that groups faces with similar facial attributes and highlights where the model is failing. A sampling method then selects faces where the model is not performing well, which are used to fine-tune the CNN. This is shown to improve driver drowsiness detection for the under-represented ethnicity groups. A representative example for non-GAN approaches is by Wang et al. \cite{wang2020towards} who studied the  mitigation strategies of data balancing, fairness through blindness, and fairness through awareness, and demonstrated that fairness through awareness provided the best results for smiling/not-smiling classification on the CelebA dataset.

%

\section{Methodology}
\label{sect:method} 

\begin{figure}[t]
\begin{minipage}[b]{0.46\textwidth}
\begin{subfigure}[b]{\linewidth}
\includegraphics[width=\linewidth]{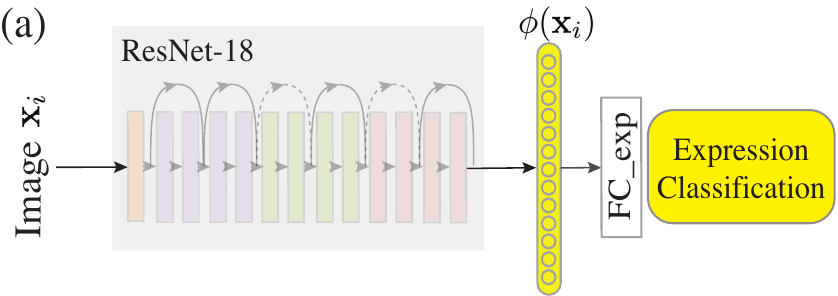}
\end{subfigure}

\vspace*{3mm} 
\begin{subfigure}[b]{\linewidth}
\includegraphics[width=\linewidth]{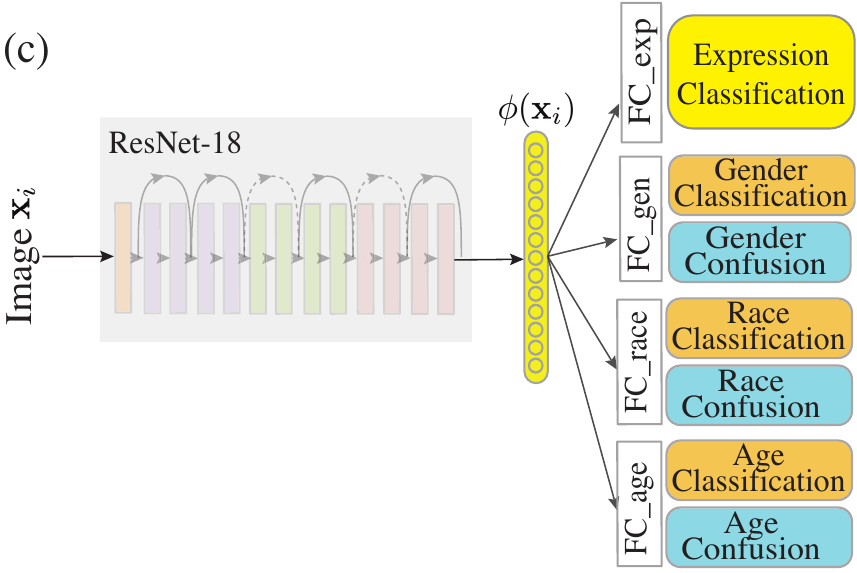}
\end{subfigure}
\end{minipage}
\hspace*{2mm}  
\begin{subfigure}[b]{0.46\textwidth}
\includegraphics[width=\linewidth]{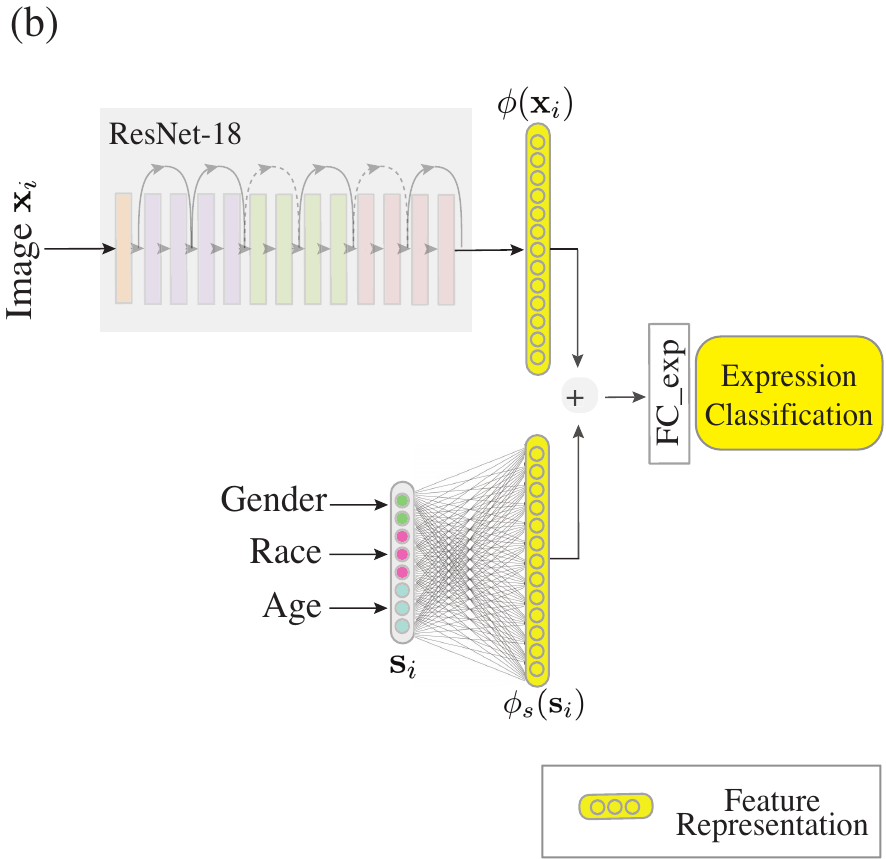}
\vspace*{1mm}
\end{subfigure}

\caption{An illustration of the three approaches. (a) The Baseline Approach, (b) The Attribute-aware Approach, (c) The Disentangled Approach}
\label{fig:architecture}
\end{figure}

To investigate whether bias is a problem for the facial expression recognition task, we conduct a comparative study using three different approaches. The first one acts as the baseline approach and we employ two other approaches, namely the Attribute-aware Approach and the Disentangled Approach, to investigate different strategies for mitigating bias. These approaches are illustrated in detail in Fig.~\ref{fig:architecture}.

\subsection{Problem Definition and Notation}
We are provided with a facial image $\mathbf{x}_i$ with a target label $y_i$. Moreover, each $\mathbf{x}_i$ is associated with sensitive labels $\mathbf{s}_i = <s_1, ..., s_m>$, where each label $s_j$ is a member of an attribute group, $s_j \in \mathcal{S}_j$. For our problem, we consider $m=3$ attribute groups (race, gender, age) , and $\mathcal{S}_j$ can be illustratively defined to be  $\{\textrm{Caucasian, African-American,}$\\$ \textrm{Asian}\}$ for $j=\textrm{race}$. The goal then is to model $p(y_i | \mathbf{x}_i)$ without being affected by $\mathbf{s}_i$.

\subsection{The Baseline Approach}
Our baseline is a Residual Network (ResNet) \cite{he2016deep}, a widely used architecture which achieved high performance for many automatic recognition tasks. We utilise a 18-layer version (ResNet-18) for our analyses. We train this baseline network with a Cross Entropy loss to predict a single expression label $y_i$ for each input $\mathbf{x}_i$:
\begin{equation}  
\mathcal{L}_{exp}(\mathbf{x}_i)=-\sum_{k=1}^{K}\mathds{1}[y_i=k]\log p_k , \label{eqn:affect_loss}
\end{equation}
where $p_k$ is the predicted probability for $\mathbf{x}_i$ being assigned to class $k\in K$; and $\mathds{1}[\cdot]$ is the indicator function.

\subsection{The Attribute-aware Approach}
Inspired by the work described in \cite{guo2010human,dwork2012fairness}, we propose an alternative ``fairness through awareness'' approach. In \cite{guo2010human,dwork2012fairness}, separate networks or branches are trained for each sensitive attribute, which is computationally more expensive. In our attribute-aware solution, we provide a representation of the attributes as another input to the classification layer (Fig. \ref{fig:architecture}(b)). Note that this approach allows us to investigate how explicitly providing the attribute information can affect the expression recognition performance and whether it can mitigate bias.

To be comparable with the baseline approach, ResNet-18 is used as the backbone network for extracting a feature vector $\phi(\mathbf{x}_i)$ from image $\mathbf{x}_i$. In order to match the size of $\phi(\mathbf{x}_i)$, which is 512 in the case of ResNet-18, the attribute vector $\mathbf{s}_i$ is upsampled through a fully-connected layer: $\phi_s(\mathbf{s}_i) = \mathbf{W}_s \mathbf{s}_i+\mathbf{b}_s$. Then, the addition $\phi_s(\mathbf{s}_i)+\phi(\mathbf{x}_i)$ is provided as input to the classification layer (Figure \ref{fig:architecture}). The network is trained using the Cross Entropy loss in Equation \ref{eqn:affect_loss}. 

\subsection{The Disentangled Approach}
The main idea for this approach is to make sure the learned representation $\phi(\mathbf{x}_i)$ does not contain any information about the sensitive attributes $\mathbf{s}_i$; in other words, we cannot predict $\mathbf{s}_i$ from $\mathbf{x}_i$ while being able to predict the target label $y_i$. To do that, we utilise the disentangling approach described in \cite{liu2018exploring,alvi2018turning}. 

For this purpose, we first extract $\phi(\mathbf{x}_i)$ using ResNet-18 to be consistent with the first two approaches. Then we split the network into two sets of branches: One primary branch for the primary classification task (expression recognition) with the objective outlined in Equation \ref{eqn:affect_loss}, and parallel branches designed to ensure that $\phi(\mathbf{x}_i)$ cannot predict $\mathbf{s}_i$. 

The parallel branches use a so-called \emph{confusion} loss to make sure that sensitive attributes cannot be predicted from $\phi(\mathbf{x}_i)$:
\begin{equation}  
\mathcal{L}_{conf}=-\sum_{\mathcal{S}_j\in\mathcal{S}} \sum_{s\in \mathcal{S}_j}\frac{1}{|\mathcal{S}_j|}\log p_{s}, \label{eqn:conf_loss}
\end{equation}
which essentially tries to estimate equal probability ($1/|\mathcal{S}_j|$) for each sensitive attribute. If this objective is satisfied, then we can ensure that $\phi(\mathbf{x}_i)$ cannot predict $\mathbf{s}_i$. However, the network can easily learn a trivial solution to not map $\phi(\mathbf{x}_i)$ to $\mathcal{S}$ even when $\phi(\mathbf{x}_i)$ contains sensitive information. To avoid this trivial solution, an attribute predictive Cross Entropy loss is also used \cite{liu2018exploring}:
\begin{equation}  
\mathcal{L}_{s}=
-\sum_{\mathcal{S}_j\in\mathcal{S}} \sum_{s\in \mathcal{S}_j} \mathds{1}[y_{s}=s]\log p_{s},
\end{equation}
which functions as an adversary to $\mathcal{L}_{conf}$ in Equation \ref{eqn:conf_loss}.

These tasks share all the layers till the final fully-connected layer $\phi()$. At the final fully-connected layer parallel branches are created for specific tasks (Figure \ref{fig:architecture}(c)). The difference between the training of the primary expression classification task and the sensitive attribute classification tasks is that the gradients of $\mathcal{L}_{s}$ are only back-propagated to $\phi()$, but do not update the preceding layers similar to \cite{liu2018exploring}.

The overall loss is then defined as:
\begin{equation}  
\mathcal{L}=\mathcal{L}_{exp} + \mathcal{L}_s +\alpha \mathcal{L}_{conf},
\end{equation}
where $\alpha$ the contribution of the confusion loss.

By jointly minimizing $\mathcal{L}_{exp}$, $\mathcal{L}_{s}$ and $\mathcal{L}_{conf}$, the final shared feature representation $\phi$ is forced to distill the facial expression information and dispel the sensitive attribute information. Alvi et al. \cite{alvi2018turning} claim that this approach can improve the classification performance when faced with an extreme bias.

\section{Implementation Details}
In this section, we provide details about how we evaluate and compare the three methods in terms of their performance for expression recognition and fairness. 

\subsection{Datasets}
Majority of the affect / expression datasets do not contain gender, age and ethnicity labels. Therefore, as a proof of concept we conduct our investigations on two well-known datasets, RAF-DB \cite{li2017reliable} and CelebA \cite{liu2015faceattributes}, that meet the following criteria: (1) providing labels for expressions of emotions and/or affect; (2) providing labels for gender, age and/or ethnicity attributes for each sample; and (3) being large enough to enable the training and evaluation of the state-of-the-art deep learning models.

{\bf RAF-DB} \cite{li2017reliable} is a real-world dataset, with diverse facial images collected from the Internet. The images are manually annotated with expression labels and attribute labels (i.e. race, age and gender of the subjects). 
For our experiments, we chose a subset of the dataset with basic emotion labels -  14,388 images, with 11,512 samples used for training and 2,876 samples used for testing. The task we focus on using this dataset is to recognize the seven categories of facial expressions of emotion (i.e. Anger, Disgust, Fear, Happiness, Sadness, Surprise and Neutral).

{\bf CelebA} \cite{liu2015faceattributes} is a large-scale and diverse face attribute dataset, which contains 202,599 images of 10,177 identities. There are 40 types of attribute annotations in the dataset. Three attributes are chosen in our experiments, including ``Smiling”, ``Male" and ``Young", corresponding to ``Facial Expression", ``Gender" and ``Age" information.
Although CelebA does not contain full range of expression annotations, it provides ``Gender" and ``Age" information that can be utilized to investigate bias and fairness. To the best of our knowledge, there is no other large-scale real-world dataset that can be used for this purpose. Thus, we conduct additional experiments on CelebA as a supplement.
Officially, CelebA is partitioned into three parts (training set, validation set and testing set). Our models are trained using the training set and evaluated on the testing set. 
The task we focus on using this dataset is to recognize the expression of “Smiling”.

\subsection{Image Pre-processing,  Augmentation and Implementation}
For both RAF-DB and CelebA datasets, images are first aligned and cropped, so that all faces appear in approximately similar positions, and normalized to a size of 100 $\times$ 100 pixels. These images are fed into the networks as input. 

Deep networks require large amounts of training data to ensure generalization for a given classification task. However, most facial expression datasets available for research purposes do not contain sufficient number of samples to appropriately train of a model, and this may result in overfitting. Therefore, we use data augmentation in this study, which is a commonly employed strategy in many recognition studies when training a network. During the training step, two strategies of image augmentation are applied to the input. For strategy one, the input samples are randomly cropped to a slightly smaller size (i.e. 96 $\times$ 96); randomly rotated with a small angle (i.e. range from -15\degree to 15\degree); and horizontally mirrored in a randomized manner. For strategy two, histogram equalization is applied to increase the global contrast of the images. Following the suggestion from \cite{kuo2018compact}, we adopt a weighted summation approach to take advantage of both strategy one and two.

All three methods are implemented using PyTorch \cite{paszke2017automatic} and trained with the Adam optimizer \cite{kingma2014adam}, with a mini-batch size of 64, and an initial learning rate of 0.001. The learning rate decays linearly by a factor of 0.1 every 40 epochs for RAF-DB and every 2 epochs for CelebA. The maximum training epochs is 200, but early stopping is applied if the accuracy does not increase after 30 epochs for RAF-DB and 5 epochs for CelebA.

\subsection{Evaluation Metrics}
When performing a classification task, the most commonly used metrics are accuracy, precision and recall, with most studies reporting comparative results using the accuracy metric. 
However, these metrics are not sufficient in exposing differences in performance (bias) in terms of  gender, age and ethnicity attributes. Therefore, we propose to evaluate the three algorithms introduced using two evaluation metrics: Accuracy and Fairness.

{\bf Accuracy} is simply the fraction of the predictions that the model correctly predicts. 
{\bf Fairness} indicates whether a classifier is fair to the sensitive attributes of gender, age and ethnicity. There are various definitions of fairness \cite{verma2018fairness}. For this study, we use the Fairness of ``equal opportunity" as described in \cite{hardt2016equality}. The main idea is that the classifier should ideally provide similar results across different demographic groups.

Let $\mathbf{x}$, $y$ be the variables denoting the input and the ground truth label targeted by the classifier; let $\hat{y}$ be the predicted variable; and let $s\in \mathcal{S}_i$ be a sensitive attribute (e.g. $\mathcal{S}_i=\{\textrm{male, female}\}$). Suppose there are $C$ classes: $c=1, ..., C$, and let $p{(\hat{y} = c | \mathbf{x})}$ denote the probability that the predicted class is $c$. Then ``equal opportunity" measures the difference between $p{(\hat{y}=c | y=c, s=s_0, \mathbf{x})}$ and $p{(\hat{y}=c | y=c, s=s_1, \mathbf{x})}$, where $s_0$, $s_1$ are e.g. male and female attributes.
Suppose there are only two demographic groups in the sensitive attribute (e.g. `female' and `male' in for attribute `gender')  considered, then the fairness measure $\mathcal{F}$ can be defined as:
\begin{equation}  
\mathcal{F} = \min\left(\frac{\sum\limits_{c=1}^{C} p{(\hat{y}=c | y=c, s=s_0, \mathbf{x})}}{\sum\limits_{c=1}^{C} p{(\hat{y}=c | y=c, s=s_1, \mathbf{x})}}, \frac{\sum\limits_{c=1}^{C} p{(\hat{y}=c | y=c, s=s_1, \mathbf{x})}}{\sum\limits_{c=1}^{C} p{(\hat{y}=c | y=c, s=s_0, \mathbf{x})}}\right).
\end{equation}

If the sensitive attribute $s$ is not binary (i.e. $s \in \{s_0, s_1, ..., s_n\}$ with $n>1$), the fairness measure can be defined to measure the largest accuracy gap among all demographic groups. Let $d$ denote the dominant sensitive group (i.e. the group which has the highest overall per-class accuracy). This is calculated by summing up the class-wise accuracy. Then the fairness measure $\mathcal{F}$ in a non-binary case is defined as:
\begin{equation}  
\mathcal{F} = \min\left(\frac{\sum\limits_{c=1}^{C} p{(\hat{y}=c | y=c, s=s_0, \mathbf{x})}}{\sum\limits_{c=1}^{C} p{(\hat{y}=c | y=c, s=d, \mathbf{x})}}, ..., \frac{\sum\limits_{c=1}^{C} p{(\hat{y}=c | y=c, s=s_n, \mathbf{x})}}{\sum\limits_{c=1}^{C} p{(\hat{y}=c | y=c, s=d, \mathbf{x})}}\right). \label{eqn:fairness}
\end{equation}

\section{Experiments and Results}

\subsection{Experiments on RAF-DB}
We first present a bias analysis on RAF-DB and perform experiments to investigate whether it is possible to mitigate this bias through the three approaches we proposed in the Methodology section.

\subsubsection{RAF-DB Bias Analysis}
RAF-DB contains labels in terms of facial expressions of emotions (Surprise, Fear, Disgust, Happy, Sad, Anger and Neutral) and demographic attribute labels along gender, race and age. The provided attribute labels are as follows - \textit{Gender}: Male, Female, Unsure; \textit{Race}: Caucasian, African-American, Asian; and \textit{Age}: 0-3 years, 4-19 years, 20-39 years, 40-69 years, 70+ years. For simplicity, we exclude images labelled as Unsure for Gender. We performed an assessment of how images in the dataset represent the different race, age and gender categories. Table \ref{RAF_Distri} shows a detailed breakdown for the testing data. Note that the distribution of the training data are kept similar to the testing data. Looking at Table~\ref{RAF_Distri}, we observe that with 77.4\%, the vast majority of the subjects in the dataset are Caucasian, 15.5\% are Asian, and only 7.1\% are African-American. 56.3\% of the subjects are female, while 43.7\% are male. Most subjects are in the 20-39 years age category, with the 70+ years category containing the fewest images.
This confirms what we have mentioned earlier, that majority of the affect/expression datasets have been acquired without a consideration for containing images that are evenly distributed across the attributes of gender, age and ethnicity. 
Therefore the goal of our experiment is to investigate whether it is possible to mitigate this bias through the three approaches we proposed in the Methodology section: the Baseline Approach, the Attribute-aware Approach, and the Disentangled Approach.

\begin{table}[t]
    \caption{RAF-DB data distribution of the test set (Cau: Caucasian, AA: African-American, pct.: percentage).}
    \centering
    \resizebox{\linewidth}{!}{%
    \scriptsize
    \begin{tabular}{|l|r|r|r|r|r|r|r|r|r|r|r|}
    \hline
         \multirow{2}{*}{} & \multicolumn{2}{c|}{Gender} & \multicolumn{3}{c|}{Race} & \multicolumn{5}{c|}{Age} & \multirow{2}{*}{pct.} \\ \cline{2-11}
         & Male & Female & Cau & AA & Asian & 0-3 & 4-19 & 20-39 & 40-69 & 70+ & \\ \hline\hline
        Surprise & 138 & 159 & 260 & 16 & 21 & 36 & 39 & 180 & 36 & 6 & 10.3\% \\
        Fear & 43 & 36 & 61 & 5 & 13 & 3 & 7 & 50 & 16 & 3 & 2.7\% \\
        Disgust & 69 & 89 & 125 & 6 & 27 & 3 & 13 & 106 & 28 & 8 & 5.5\% \\
        Happy & 429 & 712 & 855 & 98 & 188 & 43 & 216 & 581 & 264 & 37 & 39.7\% \\
        Sad & 147 & 239 & 291 & 30 & 65 & 51 & 97 & 164 & 61 & 13 & 13.4\% \\
        Anger & 119 & 45 & 144 & 10 & 10 & 2 & 16 & 115 & 28 & 3 & 5.7\% \\
        Neutral & 312 & 339 & 489 & 39 & 123 & 20 & 83 & 458 & 68 & 22 & 22.6\% \\ \hline\hline
        pct. & 43.7\% & 56.3\% & 77.4\% & 7.1\% & 15.5\% & 5.5\% & 16.4\% & 57.5\% & 17.4\% & 3.2\% &  \\ \hline
    \end{tabular}}
    \label{RAF_Distri}
\end{table}

\begin{table}[t]
    \caption{Class-wise accuracy of the models by expression labels on RAF-DB.}
    \centering
    \begin{tabular}{|l|r|r|r|r|r|r|}
    \hline
        \multirow{2}{*}{} & \multicolumn{3}{c|}{Without Augmentation} & \multicolumn{3}{c|}{With Augmentation} \\ \cline{2-7}
         & Baseline & Attri-aware & Disentangle & Baseline & Attri-aware & Disentangle \\ \hline\hline
        Mean & 65.3\% & 66.9\% & 62.2\% & 73.8\% & \underline{74.6\%} & \textbf{74.8\%} \\ \hline
        Surprise & 75.8\% & 79.7\% & 77.0\% & \textbf{82.8\%} & \underline{82.5\%} & 81.8\% \\
        Fear & 40.5\% & 47.4\% & 40.5\% & \underline{54.4\%} & \textbf{55.7\%} & 53.8\% \\
        Disgust & 41.1\% & 41.1\% & 33.1\% & 51.6\% & \underline{53.8\%} & \textbf{54.1\%} \\
        Happy & 91.6\% & 92.7\% & 92.4\% & \textbf{93.6\%} & 92.7\% & \underline{93.3\%} \\
        Sad & 63.2\% & 64.2\% & 55.4\% & 73.1\% & \textbf{80.6\%} & \underline{77.7\%} \\
        Anger & 66.5\% & 62.8\% & 53.7\% & 73.8\% & \underline{74.4\%} & \textbf{81.0\%} \\
        Neutral & 78.7\% & 80.4\% & \underline{83.5\%} & \textbf{87.6\%} & 82.2\% & 82.1\% \\ \hline
    \end{tabular}
    \label{RAF_Acc}
\end{table}

\begin{table}[t]
    \caption{Mean class-wise accuracy of the models, broken down by attribute labels on RAF-DB (Cau: Caucasian, AA: African-American, M: Male, F: Female).}
    \centering
    \begin{tabular}{|l|r|r|r|r|r|r|}
    \hline
        \multirow{2}{*}{} & \multicolumn{3}{c|}{Without Augmentation} & \multicolumn{3}{c|}{With Augmentation} \\ \cline{2-7}
         & Baseline & Attri-aware & Disentangle & Baseline & Attri-aware & Disentangle \\ \hline\hline
        Male & 65.3\% & 67.4\% & 62.5\% & 72.3\% & \underline{73.7\%} & \textbf{74.2\%} \\
        Female & 63.5\% & 64.9\% & 61.0\% & \underline{74.1\%} & \underline{74.1\%} & \textbf{74.4\%} \\ \hline\hline
        Cau & 65.9\% & 68.3\% & 63.4\% & 74.7\% & \underline{74.9\%} & \textbf{75.6\%} \\
        AA & 68.1\% & 62.8\% & 58.4\% & \underline{76.3\%} & \underline{76.3\%} & \textbf{76.6\%} \\
        Asian & 60.0\% & 59.8\% & 54.4\% & 67.8\% & \underline{69.9\%} & \textbf{70.4\%} \\ \hline\hline
        0-3 & 63.6\% & 59.9\% & 56.7\% & \textbf{80.2\%} & \underline{71.9\%} & 65.0\% \\
        4-19 & 59.5\% & 58.8\% & 57.0\% & 61.1\% & \underline{63.7\%} & \textbf{69.9\%} \\
        20-39 & 65.9\% & 68.2\% & 62.9\% & 74.9\% & \underline{75.8\%} & \textbf{76.4\%} \\
        40-69 & 65.0\% & 63.4\% & 60.1\% & \underline{73.8\%} & \textbf{74.4\%} & 72.1\% \\
        70+ & 51.3\% & 53.6\% & 51.6\% & \underline{60.8\%} & 54.3\% & \textbf{62.2\%} \\ \hline\hline
        M-Cau & 65.3\% & 69.3\% & 63.6\% & 73.3\% & \underline{73.9\%} & \textbf{74.5\%} \\
        M-AA & 77.0\% & 70.4\% & 63.2\% & 66.4\% & \textbf{80.2\%} & \underline{78.7\%} \\
        M-Asian & 61.2\% & 58.6\% & 56.2\% & 67.8\% & \underline{68.4\%} & \textbf{70.2\%} \\
        F-Cau & 64.1\% & 66.2\% & 62.2\% & 74.7\% & \underline{74.9\%} & \textbf{75.5\%} \\
        F-AA & 61.6\% & 57.9\% & 62.8\% & \textbf{87.6\%} & \underline{75.8\%} & 74.6\% \\
        F-Asian & 59.1\% & 59.5\% & 52.4\% & 65.6\% & \underline{68.4\%} & \textbf{69.0\%} \\ \hline
    \end{tabular}
    \label{RAF_classwise}
\end{table}

\begin{table}[t]
    \caption{Fairness measure of the models, broken down by attribute labels on RAF-DB (G-R: Joint Gender-Race groups).}
    \centering
    \begin{tabular}{|l|r|r|r|r|r|r|}
    \hline
        \multirow{2}{*}{} & \multicolumn{3}{c|}{Without Augmentation} & \multicolumn{3}{c|}{With Augmentation} \\ \cline{2-7}
         & Baseline & Attri-aware & Disentangle & Baseline & Attri-aware & Disentangle \\ \hline\hline
        Gender & 97.3\% & 96.3\% & 97.5\% & 97.6\% & \underline{99.5\%} & \textbf{99.7\%} \\
        Race & 88.1\% & 87.5\% & 85.8\% & 88.8\% & \underline{91.6\%} & \textbf{91.9\%} \\
        Age & 77.7\% & 78.6\% & \textbf{82.1\%} & 75.8\% & 71.6\% & \underline{81.4\%} \\
        G-R & 76.7\% & 82.2\% & 83.0\% & 74.8\% & \underline{85.3\%} & \textbf{87.7\%} \\ \hline
    \end{tabular}
    \label{RAF_fair}
\end{table}

\subsubsection{Expression Recognition}
For each method, we trained two versions, with and without data augmentation. The performance of all six models are presented in Table~\ref{RAF_Acc}. We observe that data augmentation 
increases the accuracy and this applies to almost all the expression categories. The baseline model with data augmentation provides the best performance, but the difference compared to the attribute-aware approach and the disentangled approach with data augmentation are minimal. When comparing the performance across all expression categories, we observe that the accuracy varies and this variation is closely associated with the number of data points available for each expression category (See Table \ref{RAF_Distri}). The expression category of ``Happiness" is classified with the highest accuracy, while the categories of ``Fear" and ``Disgust" are classified with the lowest accuracy.

The accuracy breakdown provided in  Table \ref{RAF_Acc} by expression labels cannot shed light on the performance variation of the classifiers across different demographic groups. Thus, in Table~\ref{RAF_classwise} we provide a detailed comparison of the accuracies broken down by each demographic group. To further shed light on the inter-play between the gender and race attributes, in Table \ref{RAF_classwise} we also include results for joint Gender-Race groups. Note that the accuracy presented in Table \ref{RAF_classwise} is class-wise accuracy, which refers to the accuracy for each expression category. In this way, we ensure that the weights of all categories are the same.
This enables a more accurate analysis of fairness, not affected by the over-represented classes in the dataset.
Note that there are only a few data points for certain subgroups (i.e. Age 0-3 years, Age 70+, African-American), so the results obtained for these groups are likely to be unstable.
From Table \ref{RAF_classwise}, we can see that for class-wise accuracy, the disentangled approach with data augmentation provides the best accuracy, with the attribute-aware approach being the runner-up. This suggests that both the attribute-aware and the disentangled approaches can improve the class-wise accuracy when equipped with data augmentation.

\subsubsection{Assessment of Fairness}
In order to provide a quantitative evaluation of fairness for the sensitive attributes of age, gender and race, we report the estimated fairness measures (obtained using Equation \ref{eqn:fairness}) for the three models in Table \ref{RAF_fair}. We observe that, compared to the baseline model, both the attribute-aware and the disentangled approaches demonstrate a great potential for mitigating bias for the unevenly distributed subgroups such as Age and Joint Gender-Race. We note that, the observed effect is not as pronounced when the distribution across (sub)groups is more or less even (e.g., results for gender, which has less gap in comparison -- Table \ref{RAF_Distri}).
For the baseline model, applying data augmentation improves the accuracy by approximately 7\% (Table \ref{RAF_Acc}), but this alone can not mitigate the bias effect. Instead, both the attribute-aware and the disentangled approaches when equipped with data augmentation achieve further improvements in terms of fairness. We conclude that, among the three approaches compared, the disentangled approach is the best one for mitigating demographic bias.

\subsubsection{Discussion}
We provided a comprehensive analysis on the performance of the three approaches on RAF-DB in Tables~\ref{RAF_Acc} - \ref{RAF_fair}. We observed that although the accuracies achieved by the three models are relatively similar, their abilities in mitigating  bias are notably different. The attribute-aware approach utilizes the attribute information to allow the model to classify the expressions according to the subjects sharing similar attributes, rather than drawing information from the whole dataset, which may be misleading. The drawback of this approach is that it requires explicit attribute information apriori (e.g., the age, gender and race of the subject whose expression is being classified)  which may not be easy to acquire in real-world applications. Instead, the disentangled approach mitigates bias by enforcing the model to learn a representation that is indistinguishable for different subgroups and does not require attribute labels at test time. This approach is therefore more appropriate and efficient when considering usage in real-world settings. In \cite{HowardEtAl-2017}, facial expressions of Fear and Disgust have been reported to be less well-recognized than those of other basic emotions  and Fear has been reported to have a significantly lower recognition rate than the other classes. Despite the improvements brought along with augmentation and the attribute-aware and disentangled approaches, we observe in Table~\ref{RAF_Acc} similar results for RAF-DB.

To further investigate the performance of all six models, we present a number of challenging cases in Table \ref{RAF_failure}. We observe that many of these have ambiguities in terms of the expression category they have been labelled with. Essentially, learning the more complex or ambiguous facial expressions that may need to be analysed along/with multiple class-labels, remains an issue for all six models. This also relates to the fact that despite its wide usage, the theory of six-seven basic emotions is known to be problematic in its inability to explain the full range of facial expressions displayed in real-world settings \cite{gunes2013categorical}.

\begin{table}[t]
    \caption{Results of a number of challenging cases from RAF-DB (Cau: Caucasian, AA: African-American, M: Male, F: Female, D.A.: Data Augmentation). Correct estimations are highlighted in bold.}
    \centering
    \resizebox{\linewidth}{!}{%
    \begin{tabular}{|c|c|c|c|c|c|c|c|c|c|}
    \hline
          \multicolumn{2}{|l|}{ } &  \includegraphics[scale=0.3, bb= 0 0 100 100]{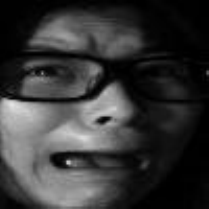} & \includegraphics[scale=0.3, bb= 0 0 100 100]{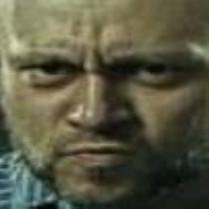} & \includegraphics[scale=0.3, bb= 0 0 100 100]{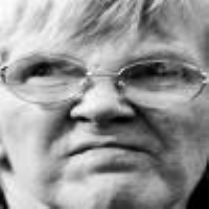} & \includegraphics[scale=0.3, bb= 0 0 100 100]{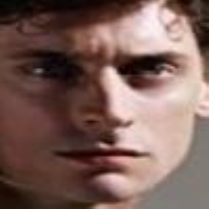} & \includegraphics[scale=0.3, bb= 0 0 100 100]{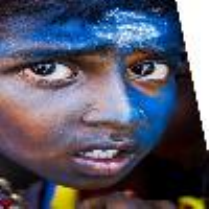} & \includegraphics[scale=0.3, bb= 0 0 100 100]{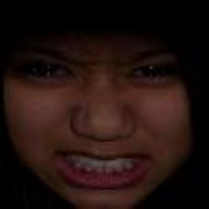} & \includegraphics[scale=0.3, bb= 0 0 100 100]{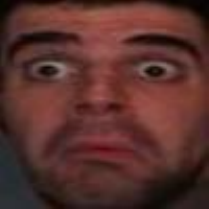} & \includegraphics[scale=0.3, bb= 0 0 100 100]{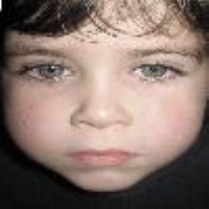} \\ \hline
         \multicolumn{2}{|r|}{Gender} & Female & Male & Female & Male & Female & Female & Male & Female \\
         \multicolumn{2}{|r|}{Race} & Asian & Cau & Cau & Cau & AA & Asian & Cau & Cau \\
         \multicolumn{2}{|r|}{Age} & 20-39 & 40-69 & 70+ & 20-39 & 4-19 & 20-39 & 40-69 & 0-3 \\ \hline\hline
         \multicolumn{2}{|r|}{Actual Label} & Fear & Anger & Disgust & Neutral & Fear & Anger & Surprise & Sad \\ \hline\hline
        \multirow{3}{*}{w/o D.A.} & Baseline & \textcolor{black}{Sad} & \textbf{Anger} & \textcolor{black}{Anger} & \textbf{Neutral} & \textcolor{black}{Neutral} & \textcolor{black}{Sad} & \textbf{Surprise} & \textbf{Sad} \\
         & Attri. & \textbf{Fear} & \textbf{Anger} & \textcolor{black}{Sad} & \textbf{Neutral} & \textcolor{black}{Happy} & \textcolor{black}{Happy} & \textbf{Surprise} & \textbf{Sad} \\
         & Disen. & \textcolor{black}{Happy} & \textcolor{black}{Disgust} & \textcolor{black}{Anger} & \textbf{Neutral} & \textcolor{black}{Surprise} & \textcolor{black}{Happy} & \textbf{Surprise} & \textbf{Sad} \\ \hline
        \multirow{3}{*}{w D.A.} & Baseline & \textbf{Fear} & \textbf{Anger} & \textbf{Disgust} & \textbf{Neutral} & \textcolor{black}{Disgust} & \textcolor{black}{Happy} & \textcolor{black}{Fear} & \textcolor{black}{Neutral} \\
         & Attri. & \textbf{Fear} & \textbf{Anger} & \textbf{Disgust} & \textcolor{black}{Sad} & \textbf{Fear} & \textcolor{black}{Happy} & \textcolor{black}{Fear} & \textbf{Sad} \\
         & Disen. & \textcolor{black}{Sad} & \textcolor{black}{Disgust} & \textcolor{black}{Anger} & \textcolor{black}{Sad} & \textbf{Fear} & \textbf{Anger} & \textbf{Surprise} & \textbf{Sad} \\ \hline
    \end{tabular}}
    \label{RAF_failure}
\end{table}

\subsection{Experiments on CelebA Dataset}
We follow a similar structure, and first present a bias analysis on CelebA Dataset and subsequently perform experiments using the three approaches we proposed in Section \ref{sect:method}.

\subsubsection{CelebA Bias Analysis}
CelebA dataset contains 40 types of attribute annotations, but for consistency between the two studies, here we focus only on the attributes ``Smiling”, ``Male" and ``Young". 
Table \ref{CelebA_Distri} shows the test data breakdown along these three attribute labels. Compared to RAF-DB, CelebA is a much larger dataset. The target attribute ``Smiling” is well balanced, while the sensitive attributes Gender and Age are not evenly distributed.

\begin{table}[t]
    \caption{CelebA data distribution of the test set.} 
    \centering
    \begin{tabular}{|l|r|r|r|r|r|}
    \hline
        \multirow{2}{*}{} & \multicolumn{2}{c|}{Gender} & \multicolumn{2}{c|}{Age} & \multirow{2}{*}{Percentage} \\ \cline{2-5}
         & Female & Male & Old & Young &  \\ \hline\hline
        Not Smiling & 5354 & 4620 & 2358 & 7616 & 50.0\% \\
        Smiling & 6892 & 3094 & 2489 & 7497 & 50.0\% \\ \hline\hline
        Percentage & 61.4\% & 38.6\% & 24.3\% & 75.7\% &  \\ \hline
    \end{tabular}
    \label{CelebA_Distri}
\end{table}

\begin{table}[t]
    \caption{Accuracy of the models, broken down by smiling labels on CelebA dataset (NoSmi: Not Smiling)}
    \centering
    \begin{tabular}{|l|r|r|r|r|r|r|}
    \hline
        \multirow{2}{*}{} & \multicolumn{3}{c|}{Without Augmentation} & \multicolumn{3}{c|}{With Augmentation} \\ \cline{2-7}
         & Baseline & Attri-aware & Disentangle & Baseline & Attri-aware & Disentangle \\ \hline\hline
        Mean & \underline{93.0\%} & \textbf{93.1\%} & 92.8\% & 92.9\% & \underline{93.0\%} & 92.9\% \\ \hline
        NoSmi & 93.9\% & \textbf{94.1\%} & \textbf{94.1\%} & 93.9\% & 93.7\% & 93.7\% \\
        Smiling & 92.1\% & 92.1\% & 91.6\% & 91.9\% & \textbf{92.2\%} & \textbf{92.2\%} \\ \hline
    \end{tabular}
    \label{CelebA_Acc}
\end{table}

\begin{table}[t]
    \caption{Mean class-wise accuracy of the models, broken down by attribute labels on CelebA dataset (M: Male, F: Female)}
    \centering
    \begin{tabular}{|l|r|r|r|r|r|r|}
    \hline
         \multirow{2}{*}{} & \multicolumn{3}{c|}{Without Augmentation} & \multicolumn{3}{c|}{With Augmentation} \\ \cline{2-7}
         & Baseline & Attri-aware & Disentangle & Baseline & Attri-aware & Disentangle \\ \hline\hline
        Female & 93.5\% & 93.5\% & 93.4\% & \textbf{93.6\%} & \textbf{93.6\%} & \textbf{93.6\%} \\
        Male & \underline{91.8\%} & \textbf{91.9\%} & 91.6\% & 91.2\% & 91.6\% & 91.5\% \\ \hline\hline
        Old & \textbf{91.6\%} & \textbf{91.6\%} & 91.4\% & 91.0\% & 91.4\% & 91.5\% \\
        Young & 93.4\% & \textbf{93.6\%} & 93.3\% & \underline{93.5\%} & \underline{93.5\%} & 93.4\% \\ \hline\hline
        F-Old & 92.1\% & 92.0\% & 92.0\% & 92.3\% & \textbf{92.7\%} & \underline{92.5\%} \\
        F-Young & 93.7\% & \textbf{93.8\%} & 93.6\% & \textbf{93.8\%} & 93.7\% & 93.7\% \\
        M-old & \textbf{90.7\%} & \underline{90.6\%} & 90.4\% & 89.6\% & 90.0\% & 90.2\% \\
        M-young & \underline{92.5\%} & \textbf{92.8\%} & 92.3\% & 92.3\% & \underline{92.5\%} & 92.3\% \\ \hline
    \end{tabular}
    \label{CelebA_classwise}
\end{table}

\begin{table}[t]
    \caption{Fairness measure of the models, broken down by attribute labels on CelebA dataset (G-A: Joint Gender-Age groups)}
    \centering
    \begin{tabular}{|l|r|r|r|r|r|r|}
    \hline
         \multirow{2}{*}{} & \multicolumn{3}{c|}{Without Augmentation} & \multicolumn{3}{c|}{With Augmentation} \\ \cline{2-7}
         & Baseline & Attri-aware & Disentangle & Baseline & Attri-aware & Disentangle \\ \hline\hline
        Gender & \underline{98.2\%} & \textbf{98.3\%} & 98.1\% & 97.4\% & 97.8\% & 97.8\% \\
        Age & \textbf{98.1\%} & 97.8\% & 97.9\% & 97.4\% & 97.8\% & \underline{98.0\%} \\
        G-A & \textbf{96.9\%} & \underline{96.6\%} & \underline{96.6\%} & 95.5\% & 96.0\% & 96.3\% \\ \hline
    \end{tabular}
    \label{CelebA_fair}
\end{table}

\subsubsection{Expression (Smiling) Recognition}
The task on this dataset is to train a binary classifier to distinguish the expression ``Smiling” from ``Non-Smiling”. The Baseline Approach, the Attribute-aware Approach, and the Disentangled Approach introduced in Section \ref{sect:method} are trained and tested on this task. Again, evaluation is performed with and without data augmentation and performance is reported in Table \ref{CelebA_Acc}. As this is a relatively simple task with sufficient samples, the accuracies of all six models do not show significant differences. In Table \ref{CelebA_classwise}, all of them provide comparable results  for class-wise accuracy broken down by attribute labels. The fairness measures reported in Table \ref{CelebA_fair} are also very close to one other.

\subsubsection{Discussion}
The results obtained for CelebA-DB could potentially be due to several reasons. Firstly, it is more than ten times larger than RAF-DB, thus the trained models do not benefit as much from data augmentation. Secondly, the bias mitigation approaches are more suitable in the context of an uneven attribute distribution or imbalanced number of data points for certain subgroups, which is not the case for CelebA-DB. Thirdly, the recognition task is a simple binary classification task, and therefore both accuracy and fairness results are already high with little to no potential for improvement. 
In light of these, presenting visual examples of failure cases does not prove meaningful and does not provide further insights. We did however manually inspect some of the results and observed that the binary labelling strategy may have introduced ambiguities. In general, when labelling affective constructs, using a continuous dimensional approach (e.g., labelling Smiling using a Likert scale or continuous values in the range of [-1,+1]) is known to be more appropriate in capturing the full range of the expression displayed \cite{gunes2013categorical}.

\section{Conclusion}
To date, there exist a large variety and number of datasets for facial expression recognition tasks \cite{LiDeng-TAC2020,Martinez2019Automatic}. However, virtually none of these datasets have been acquired with consideration of containing images and videos that are evenly distributed across the human population in terms of sensitive attributes such as gender, age and ethnicity. 
Therefore, in this paper, we first focused on quantifying how these attributes are distributed across facial expression  datasets, and what effect this may have on the performance of the resulting classifiers trained on these datasets. 
%
%
%
Furthermore, in order to investigate whether bias is a problem for facial expression recognition, we conducted a comparative study using three different approaches, namely a baseline, an attribute-aware and a disentangled approach, under two conditions w/ and w/o data augmentation. As a proof of concept we conducted extensive experiments on two well-known datasets, RAF-DB and CelebA, that contain labels for the sensitive attributes of gender, age and/or race.

The bias analysis undertaken for RAF-DB showed that the vast majority of the subjects are Caucasian and most are in the 20-39 years age category. The experimental results suggested that data augmentation improves the accuracy of the baseline model, but this alone is unable to mitigate the bias effect. Both the attribute-aware and the disentangled approach equipped with data augmentation perform better than the baseline approach in terms of accuracy and fairness, and the disentangled approach is the best for mitigating demographic bias. The experiments conducted on the CelebA-DB show that the models employed do not show significant difference in terms of neither accuracy nor fairness. Data augmentation does not contribute much as this is already a large dataset. We therefore conclude that bias mitigation strategies might be more suitable in the existence of uneven attribute distribution or imbalanced number of subgroup data, and in the context of more complex recognition tasks. 

We note that the results obtained and the conclusions reached in all facial bias studies are both data and model-driven. Therefore, the study presented in this paper should be expanded by employing other relevant facial expression and affect datasets and machine learning models to fully determine the veracity of the findings. Utilising inter-relations between other attributes and gender, age and race, as has been done by
\cite{wang2020towards}, or employing generative counterfactual face attribute augmentation and investigating its impact on the classifier output, as undertaken in \cite{Denton2019DetectingBW}, might be also able to expose other more implicit types of bias encoded in a dataset. However, this requires the research community to invest effort in creating facial expression datasets with explicit labels regarding these attributes.

\section{Acknowledgments}
The work of Tian Xu and Hatice Gunes is funded by the European Union's Horizon 2020 research and innovation programme, under grant agreement No. 826232. Sinan Kalkan is supported by Scientific and Technological Research Council of Turkey (T\"UB\.ITAK) through BIDEB 2219 International Postdoctoral Research Scholarship Program.
%
%
\bibliographystyle{splncs04}
\bibliography{ref}
\end{document}